\documentclass[10pt,twocolumn,letterpaper]{article}

\usepackage{wacv}
\usepackage{times}
\usepackage{epsfig}
\usepackage{graphicx}
\usepackage{amsmath}
\usepackage{amssymb}

\usepackage{graphicx}
\usepackage{amsmath}
\usepackage{amssymb}
\graphicspath{{images/}}
\usepackage{mathtools}
\usepackage{adjustbox}
\usepackage{caption}
\usepackage{url}
\usepackage{subcaption}

\wacvfinalcopy 

\def\wacvPaperID{424} 

\ifwacvfinal\fi
\begin{document}

\title{ReHAR: Robust and Efficient Human Activity Recognition}

\author{Xin Li \hspace{25mm}  Mooi Choo Chuah\\
Department of Computer Science and Engineering, Lehigh University\\
{\tt\small xil915@lehigh.edu \hspace{5mm} chuah@cse.lehigh.edu}
}

\maketitle
\ifwacvfinal\thispagestyle{empty}\fi

\begin{abstract}
Designing a scheme that can achieve a good performance in predicting single person activities and group activities is a challenging task. In this paper, we propose a novel robust and efficient human activity recognition scheme called ReHAR, which can be used to handle single person activities and group activities prediction. First, we generate an optical flow image for each video frame. Then, both video frames and their corresponding optical flow images are fed into a Single Frame Representation Model to generate representations. Finally, an LSTM is used to predict the final activities based on the generated representations. The whole model is trained end-to-end to allow meaningful representations to be generated for the final activity recognition. We evaluate ReHAR using two well-known datasets: the NCAA Basketball Dataset and the UCFSports Action Dataset. The experimental results show that the proposed ReHAR achieves a higher activity recognition accuracy with an order of magnitude shorter computation time compared to the state-of-the-art methods.   
  
\end{abstract}
\section{Introduction}
\label{sec:intro}
With technology advancement in embedded system design, powerful cameras have been embedded within smartphones, and wireless cameras can be easily deployed at street corners, traffic lights, big stadiums, train stations, etc.  In addition, the growth of online media, surveillance and mobile cameras have resulted in explosion of videos being uploaded to social media sites such as Facebook, Youtube. For example, it is reported that over 300 hours of video are uploaded every minute to Youtube servers. The availability of such vast volume of videos has attracted the computer vision community to conduct much research on human activity recognition since people are arguably the most interesting subjects of such videos. Automatic human activity recognition allows engineers and computer scientists to design smarter surveillance systems, semantically aware video indexes and also more natural human computer interfaces. 

Law enforcement tasked with monitoring a large crowd event, e.g., Macy’s Thanksgiving parade will appreciate having quick human activity recognition analysis on tons of videos captured by cameras deployed over the streets to quickly identify suspicious or criminal behaviors. Similarly, sport fans who may not be able to watch big games in real time will be thrilled if TV broadcasters can provide video-based sport highlights which they can enjoy without watching the whole 3-4 hours games.  
Image or video based social media sites such as Youtubes are also interested in automatically classifying millions of uploaded videos to provide semantic-aware video indexes to facilitate easy search by viewers. Furthermore, drones have been deployed to conduct surveillance after big natural disaster events, e.g., hurricanes. Having efficient human activity recognition from real-time videos allows emergency workers to quickly spot a small group of people waving on top of the roofs waiting to be rescued or criminals attempting to loot shops for goods while others' attentions are focused on rescue missions.

Despite the explosion of video data, the ability to automatically recognize and understand human activities is still rather limited. This is primarily due to multiple challenges inherent to the recognition task, namely large variability in human execution styles, complexity of the visual stimuli in terms of camera motion, background clutter, viewpoint changes, etc, and the number of activities that can be recognized. Much recent work has proposed deep architectures for activity recognition \cite{baccouche10, baccouche11, donahue2015long, karpathy14, simonyan14}. \cite{karpathy14, simonyan14} both propose convolutional networks which learn filters based on a stack of $N$ input frames but such fixed length approaches cannot learn to recognize complex video sequences, e.g., cooking sequences as presented in \cite{das13}. \cite{baccouche10} uses recurrent neural networks to learn temporal dynamics using traditional vision feature \cite{baccouche10} while \cite{baccouche11} uses deep features but both do not train their models end--to-end and hence may not perform well on more complex video sequences. A handcrafted video representation capturing short, medium, and long action dynamcis has also been proposed \cite{licvpr16}. However, these approaches cannot infer group activities such as those found in sport related videos captured during volleyballs or basketball tournaments. Recent works have proposed hierarchical-based LSTM models \cite{msibrahimCVPR16, feifei16} for group activity recognition but these approaches typically consume huge cloud resources and often run slowly. Faster schemes need to be designed for there are application scenarios that mandate real time requirements, e.g., sport highlights in big games. Computation time becomes more important when we run a deep learning model on mobile devices \cite{li2017deeprebirth, li2016deepcham}. 

In this paper, we propose a robust and efficient human activity recognition scheme, ReHAR, that can infer complex human activities from trimmed video clips and is trainable end-to-end.

In summary, our contributions of this paper include:
\begin{itemize}
\item design a robust and efficient human activity recognition scheme to recognize complex human activities, e.g., group activities in sport games. 
\item extensive evaluation using two popular activity datasets show that our scheme achieves higher accuracy and runs an order of magnitude faster than existing schemes.
\item explore the visual explanation for our model to understand what it has learned.
\end{itemize}

The rest of this paper is organized as follows. In Section \ref{sec:related_work}, we briefly discuss related work, followed by the introduction of some important building blocks in Section \ref{sec:building_blocks}. In Section \ref{sec:proposed_solution}, we describe our proposed activity recognition scheme and implementation details. We report our experimental results in Section \ref{sec:experiments}. Finally, we conclude this paper in Section \ref{sec:conclusion}.
\section{Related work}
\label{sec:related_work}
Much work has been done on activity recognition so here we merely summarize the more recent work on group activity recognition which many existing activity recognition schemes cannot handle.

In recent years, researchers have started to work on group activity recognition. Most existing work on group activity recognition has used hand-crafted features in structured models to represent information between individuals in space and time domains \cite{lancvpr12, lan2012discriminative,ramanathan13}. All these approaches however merely use shallow hand crated features and typically adopt a linear model that suffers from representation limitation. In \cite{msibrahimCVPR16, feifei16}, the authors propose a hierarchical model that uses a lower layer LSTM to track each individual and a higher layer LSTM that fuses information from the lower layer LSTMs to recognize group activities. Unfortunately, such approaches are computationally expensive. Thus, a more computationally efficient method must be designed to infer group activities for real time situation awareness applications. 

In \cite{li2017sbgar}, the authors proposed a semantics based group activity recognition scheme that uses an LSTM model to generate a caption for each video frame and then use another LSTM to predict the final activity categories based on generated captions. Although it achieves a higher accuracy with a shorter running time, it has at least three weaknesses: (1) the caption generation model cannot always generate a perfect caption; (2) the caption generation model is trained only based on its own loss without getting any feedback from the final output the the model. This will result in the generated captions do not contain useful information for the second model to predict the final activities; (3) it is very difficult to access a large dataset which contain caption information, e.g. individual actions in datasets used in \cite{li2017sbgar}.
\section{Important Building Blocks}
\label{sec:building_blocks}
We give a brief introduction about some building blocks before we discuss the details of our proposed scheme. 

\textbf{1. Optical Flow:} Optical Flow was first proposed by Horn et. al \cite{horn1981determining}. It is used to describe how each point in the scene moves from a frame to the next. Many improvements have been introduced \cite{brox2004high, wedel2009structure}. Recently, machine learning methods \cite{rosenbaum2013learning, leordeanu2013locally, fischer2015flownet, ilg2016flownet} have been used to estimate optical flow by taking two images as their inputs. Among all of these solutions, FlowNet 2.0 \cite{ilg2016flownet} achieved the most impressive results by using a stacked structure and fusion network. 

\textbf{2. Image Feature Extraction Via CNN:} Convolutional Neural Network (CNN) \cite{lecun1998gradient} is a type of feed-forward artificial neural network. It has been widely used in solving different types of tough tasks, e.g. natural language processing \cite{blunsom2014deep}, image recognition \cite{lawrence1997face}, etc. It has been proved that the CNN features contain more representative information of an image than other manually designed feature, e.g. SIFT, by Fischer et al. \cite{fischer2014descriptor}. Furthermore, Donahue et al. \cite{donahue2015long} used CNN as a feature extractor to recognize human activities from videos, which showed that CNNs could extract useful information related to activities. 

\textbf{3. Global Average Pooling:} In \cite{lin2013network}, Lin et al. first proposed the Global Average Pooling (GAP) layer. For a traditional image classification network, e.g. VGG16, they first changed the number of the channels in the final Max Pooling layer (block5\_pool in VGG16), so that each feature map at this layer corresponded to one image category in the dataset. Then, they replaced the Flatten and the Fully Connected (FC) layers with a GAP layer. The GAP layer took the average of each feature map and fed the results directly into a softmax layer. Based on their experimental results, using GAP layer, a model achieved a slightly better performance than using FC layers. Comparing to the FC layer, the GAP layer has at least two advantages: (1) The GAP layer enforces correspondences between feature maps and categories, thus the feature maps can be easily interpreted as categories confidence maps. (2) There is no parameter to optimize in GAP, thus overfitting is avoided at this layer. 

\begin{figure*}[ht]
	\centering
	\includegraphics[width=0.9\textwidth]{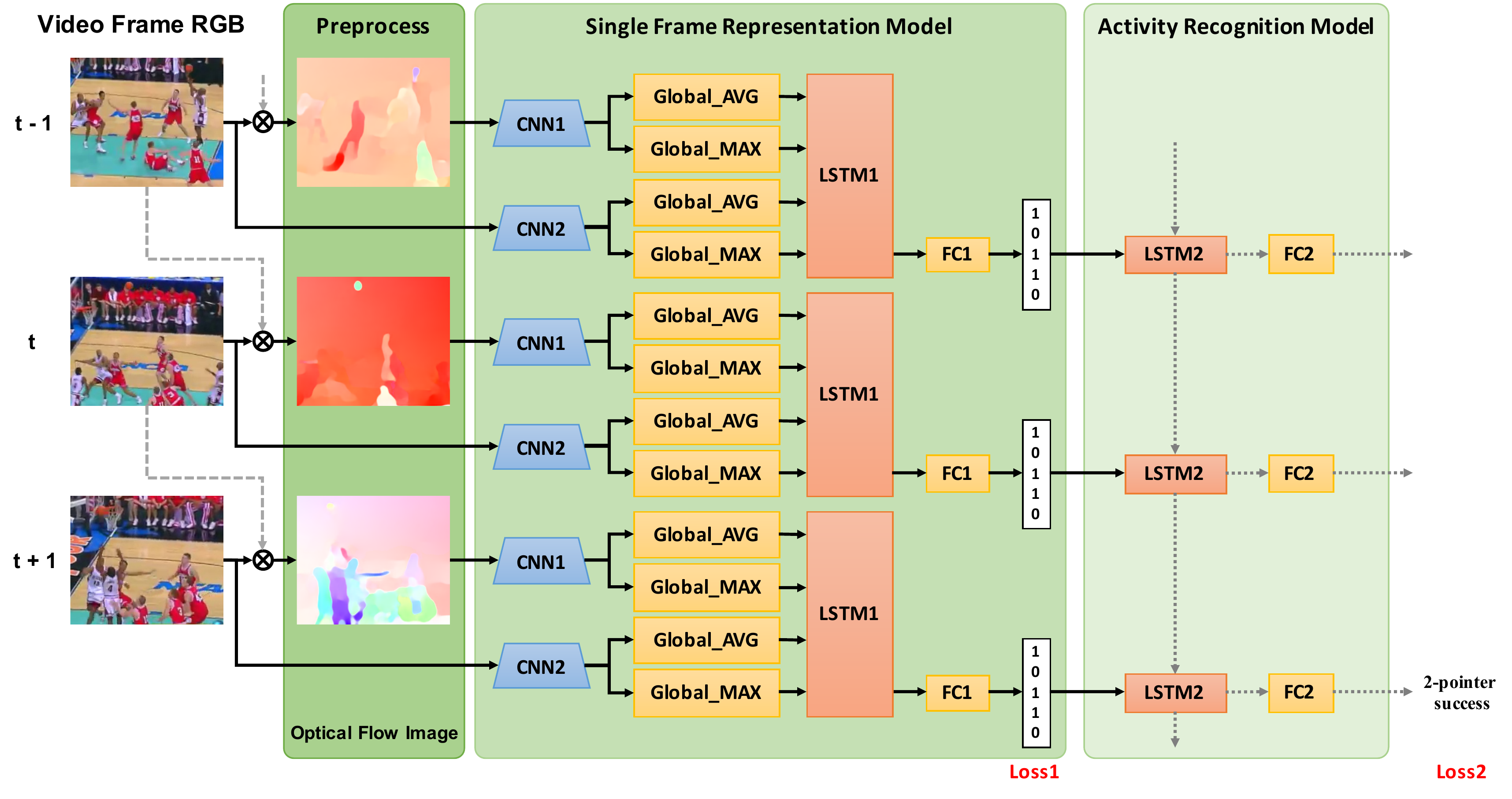}
	\vspace{-3mm}
	\caption{The architecture of the proposed Scheme. Symbol $\otimes$ indicates the operation of computing the dense optical flow using two continuous frames. CNN1 and CNN2 indicate VGG16 (layer ``block1\_conv1'' to layer ``block5\_pool'').}
	\label{fig:framework}
	\vspace{-4mm}
\end{figure*}

\textbf{4. Long Short Term Memory:} Long Short Term Memory (LSTM) model is a particular type of Recurrent Neural Network (RNN) that was first proposed by Hochreiter et al. in \cite{hochreiter1997long}. Because of its more powerful update equations and appealing back-propagation dynamics, the LSTM Network works slightly better than the traditional RNN model in practice. Using an LSTM model, Donahue et al. \cite{donahue2015long} proposed a scheme that yielded a good performance in the tasks of activity recognition, image description, and video description. Moreover, a Neural Image Caption model based on LSTM was proposed by Vinyals et al. \cite{vinyals2016show} to automatically describe the content of an image. Zhang et. al \cite{zhang2017fcn} use an LSTM model to count vehicles in city cameras. All of these works prove that the LSTM Network has the capability to extract useful information from its inputs and generate distinguishing representations. 

\section{Proposed Scheme}
\label{sec:proposed_solution}

We propose a novel end-to-end model for recognizing activities in videos. The intuition of our model is that if we can generate a good representation for every single frame, then the model will be easier to infer the final activity label for the whole video based on these representations. The model, illustrated in Figure \ref{fig:framework}, consists of three components: (1) Input Preprocessing Model, (2) Single Frame Representation Model, and (3) Activity Recognition Model. 

\subsection{Preprocessing}
We share the same position as Li and Chuah \cite{li2017sbgar} that both the background context and the motion of people contribute towards the group activity recognition. Thus, we also use the original frames (contain environment information) and their corresponding optical flow images (provide motion information) in our scheme. 
During the preprocessing phase, we feed a video frame (at time $t$) and its previous one (at time $t-1$) to the FlowNet 2.0 \cite{ilg2016flownet} to compute optical flow, since FlowNet 2.0 provides the best performance for generating optical flow. 
Then, we use the method described in \cite{baker2011database} \footnote{http://vision.middlebury.edu/flow/} to visualize the optical flow information into a colorful image (3 channels), namely optical flow image. We generate an optical flow image for every frame (except the first one) in a video. The generated optical flow images are illustrated in Figure \ref{fig:framework}.

\subsection{Single Frame Representation Model}
The Single Frame Representation Model consists of two CNN feature extractors (one for video frame and another for optical flow image) and an LSTM model. Although any CNN model can be used as a feature extractor in our model, to simplify the explanation, VGG16\footnote{VGG16: [block1\_conv1, block1\_conv2, block1\_pool, block2\_conv1, block2\_conv2, block2\_pool, block3\_conv1, block3\_conv2, block3\_conv3, block3\_pool, block4\_conv1, block4\_conv2, block4\_conv3, block4\_pool, block5\_conv1, block5\_conv2, block5\_conv3, block5\_pool, flatten, fc1, fc2, prediction]} \cite{simonyan2014very} is used in this section and the size of the video frames and the optical flow images are fixed to (224x224x3). 

Once we get the optical flow image at time $t$, we feed it as well as the corresponding video frame to two CNN models (``CNN1'' and ``CNN2'' correspondingly in Figure \ref{fig:framework}) to extract features. Instead of only removing the last prediction layer from VGG16 as in \cite{donahue2015long}, we remove the last 4 layers (flatten, fc1, fc2, and prediction). Thus, the ``CNN1'' and ``CNN2'' model in Figure \ref{fig:framework} include layers from ``block1\_conv1'' to ``block5\_pool'' of VGG16. The layer ``block5\_pool'' has (7x7x512) output size. Then, we add a Global Average Pooling layer and a Global Maximum pooling layer (``Global\_AVG'' (1x512)  and ``Global\_MAX'' (1x512) respectively in Figure \ref{fig:framework}) to its end. The advantages of doing this have been discussed in Section \ref{sec:building_blocks}. After that, we feed the output of these global pooling layers to an LSTM model (LSTM1 in Figure \ref{fig:framework}), which means the LSTM model has 4 input steps and each step has 512 dimensions. A fully-connected layer (FC1 in Figure \ref{fig:framework}) with a ``softmax'' activation function is added to the output of the final step of the LSTM1 to generate the representation for each input video frame. 

\subsection{Activity Recognition Model}
The model predicts the final activity label based on a sequence of generated single frame representations. The Activity Recognition Model is an LSTM network (LSTM2 in Figure \ref{fig:framework}) that takes the single frame representations as its input. Thus, the input step $time\_step$ of LSTM2 equals to the number of the current input video frames. In Figure \ref{fig:framework}, $time\_step = 3$. Then, the output of the final step of the LSTM2 is fed into a fully-connected layer (FC2 in Figure \ref{fig:framework}) with a ``softmax'' activation function to predict the final activity label.

\subsection{Implementation Details}
Our scheme is implemented using Python Programming Language and Keras Library \cite{chollet2015keras} with Tensorflow \cite{abadi2016tensorflow} backend. We report the implementation details of our scheme and the settings  of important parameters as follows.

\textbf{Optimization:} We train our model as a multi-task learning. The overall loss can be computed as: 
\begin{equation}
\label{equation: total_loss}
\vspace{-2mm}
Loss = (\sum\limits_{t=1}^{time\_step} loss_{1,t}) + \lambda*loss_2
\end{equation}
where $time\_step$ is the number of frames based on which the model predicts the final activity label (in Figure \ref{fig:framework}, $time\_step = 3$), $loss_{1,t}$ is the loss of the generated single frame representation at time $t$ and $loss_2$ is the loss of prediction of the final activity. $\lambda$ is the parameter that is used to balance the single frame representation generation loss and the final activity classification loss. In our experiment, we set $\lambda = 2$ to assign a higher weight to the final activity prediction, considering that the final activity prediction is our final purpose. The model is trained to minimize the $Loss$.

\textbf{Single Frame Representation Model:} The LSTM1 is a single layer LSTM with 200 hidden units. For FC1 layer, we set the dimension of its output to the number of the final activities and its training ground truth to be the one-hot vector of the final activity label. We train the Single Frame Representation Model as a classification task. In this case the representation is the probability distribution of each video frame over all activities. Thus, the $loss_1$ at time $t$, donated as $loss_{1, t}$, can be computed using categorical cross entropy loss:
\begin{equation}
\label{equation: loss1}
\vspace{-2mm}
loss_{1, t} = - \sum_{i=1} g_{t, i} \log{(p_{t, i})}
\end{equation}
where $g$ are the ground truth and $p$ are the predictions. During the testing phase, the model will generate a probability vector as a representation for each frame. 

\textbf{Activity Recognition Model:} The LSTM2 is also a single layer LSTM with 200 hidden units. The output of the FC2 is set to the number of categories. To train the model for a classification task, we train the $loss_2$ using categorical cross entropy loss:
\begin{equation}
\label{equation: loss1}
\vspace{-2mm}
loss_{2} = - \sum_{i=1} g_{i} \log{(p_{ti})}
\end{equation}
where $g$ are the ground truth and $p$ are the predictions.

\textbf{Training Process:} To speed up the training process and get a better performance, we load the pre-trained VGG16 weights on Imagenet dataset \cite{deng2009imagenet}. We train the model using ``rmsprop'' optimizer with 0.001 learning rate and 1e-8 fuzz factor until the loss becomes converged. Then, we switch the optimizer to SGD with 0.0001 learning rate. The ``rmsprop'' optimizer helps the model converge quickly, and the SGD with a small learning rate helps to tune the model.  
\section{Experiments}
\label{sec:experiments}
We run our scheme on a desktop running Ubuntu 14.04 with 4.0GHz Intel Core i7 CPU, 128GB Memory, and a NVIDIA GTX 1080 Graphics Card.

\subsection{Datasets}
We evaluate our scheme on two well known activity recognition datasets: NCAA Basketball Dataset \cite{feifei16} and UCF Sports Action Dataset \cite{rodriguez2008action}. 

\textbf{NCAA Basketball Dataset:} The NCAA Basketball Dataset\footnote{\url{http://basketballattention.appspot.com/}} was collected by Ramanathan et al. \cite{feifei16} to evaluate the performance of activity recognition schemes on multi-person action videos. It is a subset (257 Basketball Game videos) of the 296 NCAA games available from YouTube\footnote{\url{https://www.youtube.com/user/ncaaondemand}}. All videos are randomly split into 212 training, 12 validation and 33 testing videos. Each of these videos are split into 4 second clips and sub-sampled to 6fps. They filter out clips which are not profile shots, which results in a total of 11436 training, 856 validation, and 2256 testing video clips. Each of these video clips is manually labeled as one of these 11 labels: 3-pointer success, 3-pointer failure, free-throw success, free-throw failure, layup success, layup failure, other 2-pointer success, other 2-pointer failure, slam dunk success, slam dunk failure or steal success. The Basketball Dataset also annotates the bounding boxes of all the players in a subset of 9000 frames from the training videos. In our scheme, we do not use this location annotation. 

\textbf{UCF Sports Action Dataset:} The UCF Sports dataset\footnote{\url{http://crcv.ucf.edu/data/UCF_Sports_Action.php}} \cite{rodriguez2008action} consists of a set of actions collected from a wide range of stock footage websites including BBC Motion gallery and GettyImages. It consists of a total of 150 videos. Each video has one of these 10 cation categories: diving, golf swing, kicking, lifting, riding horse, running, skateboarding, swinging-bench, swinging-side, and walking.

\subsection{Metrics}
\textbf{Mean Average Precision (mAP):} Mean Average Precision is the mean of the average precision (AP) scores for each classification category. By computing a precision and recall, one can plot a precision-recall curve, plotting precision $p(r)$ as a function of recall $r$. Average precision computes the average value of $p(r)$ over the interval from $r=0$ to $r=1$ (please refer to wikipedia.org\footnote{\url{https://en.wikipedia.org/w/index.php?title=Information_retrieval}}):
\begin{equation}
AP = \int_{0}^{1} p(r)dr
\end{equation}

\textbf{Confusion Matrix:} A confusion matrix \cite{kohavi1998glossary} contains information about actual and predicted classifications generated by a classification system. In a confusion matrix, each row represents the predicted classes, while each column represents the instances of an actual class.

\subsection{Experiments on the NCAA Basketball Dataset}
In this section, we report our experimental results on the NCAA Basketball Dataset. As described in \cite{feifei16}, we classify isolated video clips into 11 classes without using any additional negative from other parts of the basketball videos. Each video clip has 24 frames (6fps for 4 seconds). The results are reported in Table \ref{table:comparison_basketball}. Among all 11 categories, our scheme achieves the highest accuracy at 8 categories compared to other baseline models. Overall, our scheme shows a 7.3\% accuracy improvement compared to \cite{feifei16} (Atten. track in Table \ref{table:comparison_basketball}). We notice that all methods perform much poorer for categories such as ``slam dunk failure''. This is because we have very little data (47 training samples and 5 testing samples) belonging to ``slam dunk failure'' category in the Basketball dataset. The performance is much better for ``free-throw'' and ``3-pointers'', because these events have fixed and more obvious patterns (especially for ``free-throw'') and more training data in this dataset. 

\begin{table*}[t!]
  \centering  
  \begin{adjustbox}{width=\textwidth}
  \begin{tabular}{|l|c|c|c|c|c|c|c|c|c|c|c|c|} 
    \hline
    & 3point S. & 3point F. & throw S. & throw F. & layup S. & layup F. & 2point S. & 2point F. & dunk S. & dunk F. & steal & Mean \\
    \hline
    IDT \cite{wang2011action} & 0.370 & 0.501 & 0.778 & 0.365 & 0.283 & 0.278 & 0.136 & 0.303 & 0.197 & 0.004 & 0.555 & 0.343 \\
    IDT \cite{wang2011action} player & 0.428 & 0.481 & 0.703 & 0.623 & 0.300 & 0.311 & 0.233 & 0.285 & 0.171 & \textbf{0.010} & 0.473 & 0.365 \\
    C3D\cite{tran2014c3d} & 0.117 & 0.282 & 0.642 & 0.319 & 0.195 & 0.185 & 0.078 & 0.254 & 0.047 & 0.004 & 0.303 & 0.221 \\
    MIL\cite{andrews2003support} & 0.237 & 0.335 & 0.597 & 0.318 & 0.257 & 0.247 & 0.224 & 0.299 & 0.112 & 0.005 & 0.843 & 0.316 \\
    LRCN\cite{donahue2015long} & 0.462 & 0.564 & 0.876 & 0.584 & 0.463 & 0.386 & 0.257 & 0.378 & 0.285 & 0.027 & 0.876 & 0.469 \\
    Atten. no track\cite{feifei16} & 0.583 & 0.668 & 0.892 & 0.671 & 0.489 & 0.426 & 0.281 & 0.442 & 0.210 & 0.006 & 0.886 & 0.505 \\
    Atten. track\cite{feifei16} & 0.600 & 0.738 & 0.882 & 0.516 & 0.500 & \textbf{0.445} & 0.341 & 0.471 & \textbf{0.291} & 0.004 & 0.893 & 0.516 \\
    \hline
    Ours & \textbf{0.753} & \textbf{0.766} & \textbf{0.933} & \textbf{0.857} & \textbf{0.613} & 0.435 & \textbf{0.405} & \textbf{0.542} & 0.232 & 0.007 & \textbf{0.940} & \textbf{0.589} \\
    \hline
  \end{tabular}
  \end{adjustbox}
  \caption{Mean average precision for event classification given isolated clips of Basketball Dataset. ``S.'' stands for ``success'' and ``F.'' stands for ``failure''. All results except ours are extracted from \cite{feifei16}.}
    \label{table:comparison_basketball}
      \vspace{-4mm}
\end{table*}

\begin{figure}[h]
	\centering
	\includegraphics[width=\linewidth]{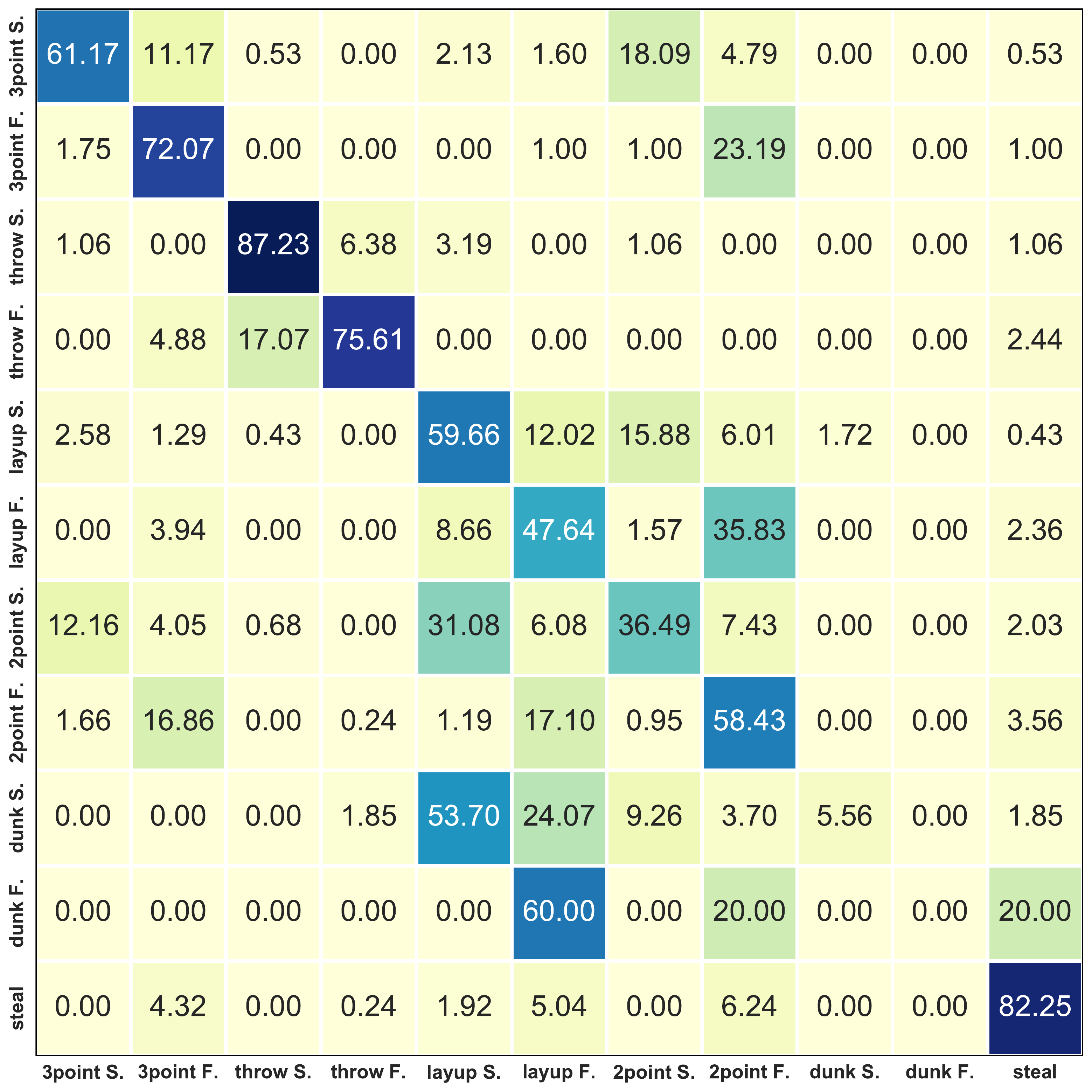}
	\caption{Confusion matrix of action recognition results on NCAA Basketball Dataset.}
	\label{fig:confusion_basketball}
	  \vspace{-4mm}
\end{figure}

The confusion matrix for all 11 actions is shown in Figure \ref{fig:confusion_basketball}. By analyzing this confusion matrix, one can see that: \textbf{(1)} 18.09\% ``3-pointer success'' test samples are incorrectly labeled as ``2-pointer success'' and 23.19\% ``3-pointer failure'' are labeled as ``2-pointer failure''. In contrast, 12.16\% and 16.86 \% ``2-pointer success/failure'' test samples are incorrectly labeled as ``3-pointer success/failure'' correspondingly. Based on the rule specification ``A player's feet must be completely behind the three-point line at the time of the shot or jump in order to make a three-point attempt; if the player's feet are on or in front of the line, it is a two-point attempt.\footnote{\url{https://en.wikipedia.org/wiki/Three-point_field_goal}}'', one can easily understand that sometime it is hard for a model (even for a person) to extract such detail information to distinguish between 3-pointers and 2-pointers. Although the authors in \cite{feifei16} designed a model to locate the ``shooter'', they still cannot extract useful enough features to achieve a better performance than our proposed scheme. \textbf{(2)} 53.7\% and 60.0\% ``slam dunk success/failure'' are predicted as ``layup success/failure''. The reason is two-fold: a. the training data for ``slam dunk success/failure'' are not enough; b. ``layup'' and ``slam dunk'' have similar action patterns (the shooter jumps under the net and sends the ball to the net).

Here, we would like to highlight an interesting observation. If we group 10 shooting-related actions (except ``steal'') into two categories (success or failure), then we will get 717 success samples and 1122 failure samples in the testing subset. Based on the output of our model, 88\% of the test samples (583 success and 1035 failure) are correctly labeled to these two categories. This observation proves that our scheme has the capability to distinguish between shooting success and shooting failure. Sometime, it is hard for people to judge if a shooting is success or not only based on the relative location between the ball and the net, let alone a designed model. Thus, we believe that our scheme achieves a good performance for it benefits from its capability to analyze players' behaviors before and after shooting, and infer the final activity label based on these behaviors. We will discuss more later by visualizing our proposed model.

\subsection{Experiments on the UCF Sports Action Dataset}
In this subsection, we evaluate the performance of our scheme using the UCF Sports Action Dataset. We follow Lan et al. \cite{lan2011discriminative} to split the dataset into training (103 videos) and testing (47 videos) subsets\footnote{\url{http://cs.stanford.edu/~taranlan/other/train_test_split}}. Among all video clips, the minimum length is 2.2 seconds and the maximum length is 14.4 seconds. We down-sampling all video clips to 24 frames before feeding them into our model. In Table \ref{table:comparison_ucfsports}, we compare our scheme to other state-of-the-art solutions. Our scheme gets the highest prediction accuracy on 8 out of 10 categories. Comparing to \cite{hou2017tube}, our scheme achieves 6.1\% accuracy improvement. In addition, we want to highlight that our scheme performs a perfect prediction (1.0 average precision) on 6 categories. This proves that our model generates more distinguishing features that benefit our model performs better than other existing methods in the task of activity recognition. 

\begin{table*}[t!]
  \centering  
  \begin{adjustbox}{width=\textwidth}
  \begin{tabular}{|l|c|c|c|c|c|c|c|c|c|c|c|} 
    \hline
    & Diving & Golf & Kicking & Lifting & Riding & Run & SkateB. & Swing & SwingB. & Walk & mAP \\
    \hline
    Gkioxari et al. \cite{gkioxari2015finding} & 0.758 & 0.693 & 0.546 & 0.991 & 0.896 & 0.549 & 0.298 & 0.887 & 0.745 & 0.447 & 0.681 \\
    Weinzaepfel et al. \cite{weinzaepfel2015learning} & 0.607 & 0.776 & 0.653 & 1.000 & 0.995 & 0.526 & 0.471 & 0.889 & 0.629 & 0.644 & 0.719 \\
    Peng et al. \cite{peng2016multi} & 0.961 & 0.805 & 0.735 & 0.992 & 0.976 & 0.824 & 0.574 & 0.836 & 0.985 & 0.760 & 0.845 \\
    Hou et al. \cite{hou2017tube} & 0.844 & 0.908 & 0.865 & 0.998 & 1.000 & \textbf{0.837} & \textbf{0.687} & 0.658 & 0.996 & 0.878 & 0.867 \\
    \hline
    Ours & \textbf{1.000} & \textbf{0.955} & \textbf{1.000} & \textbf{1.000} & \textbf{1.000} & 0.806 & 0.626 & \textbf{1.000} & \textbf{1.000} & \textbf{0.888} & \textbf{0.928} \\
    \hline
  \end{tabular}
  \end{adjustbox}
  \caption{Mean average precision for event classification given isolated clips of UCF Sports Action Dataset. All results except ours are extracted from \cite{hou2017tube}.}
    \label{table:comparison_ucfsports}
      \vspace{-4mm}
\end{table*}

\begin{figure}[h]
	\centering
	\includegraphics[width=\linewidth]{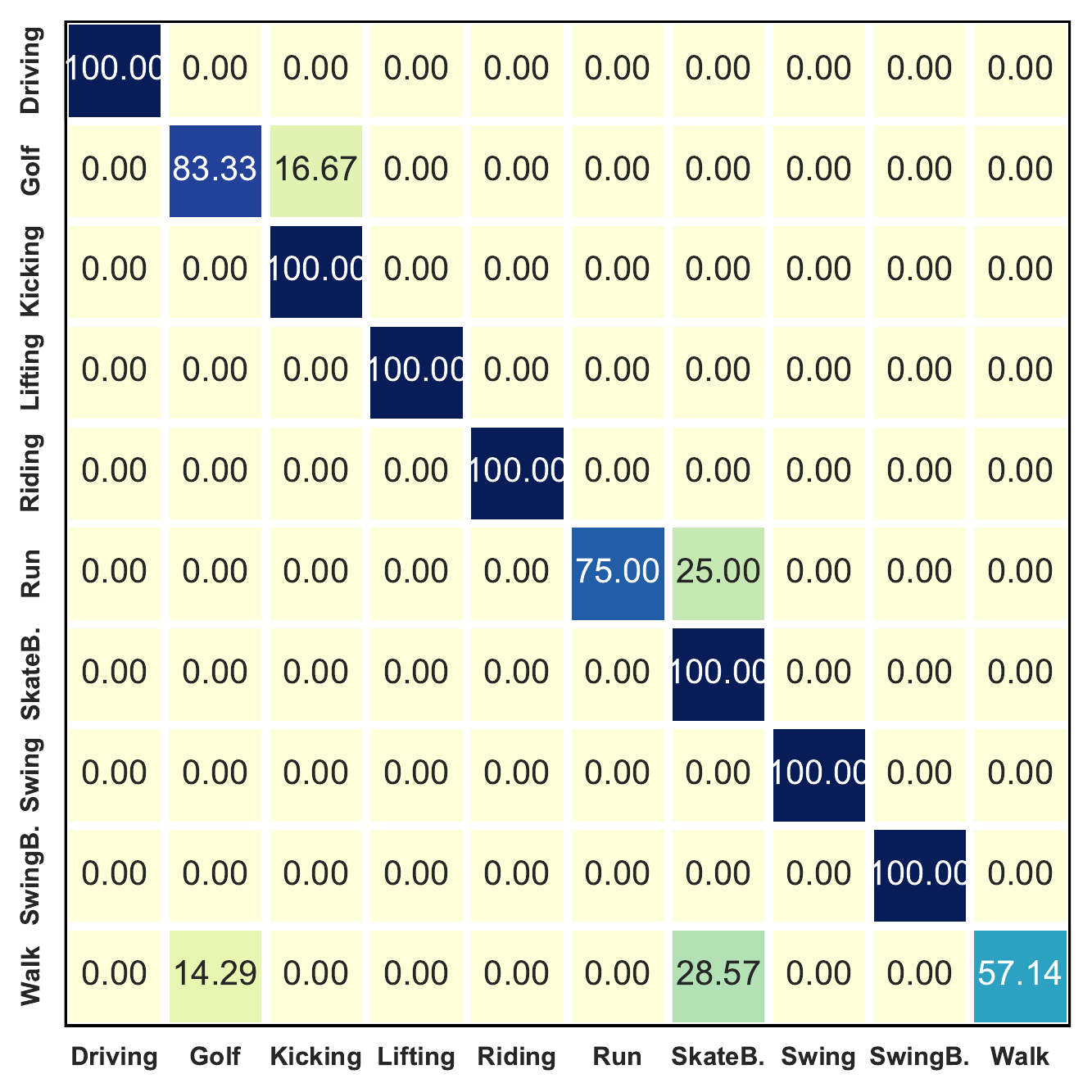}
	\caption{Confusion matrix of action recognition results on UCF Sports Action Dataset.}
	\label{fig:confusion_sports}
	  \vspace{-4mm}
\end{figure}

Please refer to Figure \ref{fig:confusion_sports} for more details about our results on the UCF Sports Dataset. One can see that our scheme performs very well on most categories. However, it incorrectly labels some ``Walking'' testing samples to ``Golf'' and ``SkateBoarding''. This is because these samples have some similiar features as samples in those incorrect categories. For example, in video ``Walk-Front/006RF1-13902\_70016.avi'', there is a person walking on a golf course with a golf pole. The environment is definitely related to golf and the motion of the golf pole looks like a person is swinging the pole in front of him. More details will be discussed later by visualizing the model.

\subsection{Computation Time}
As we have already discussed in Section \ref{sec:intro}, in some application scenarios, predicting an activity label in real time is highly important. Thus, in this subsection, we report the computation time of ReHAR. Computing optical flow images takes FlowNet 2.0 \cite{ilg2016flownet} around 7ms (140 fps). We report the computation time of ReHAR (including optical flow images generation time) using different CNN models as base net in Table \ref{table:computation_time}. In total, our model (using VGG16 as its base net) takes 103.65 ms to process 10 input frames and 239.04 ms for 24 input frames. In \cite{li2017sbgar}, the computation time of SBGAR model using InceptionV3 as feature extractor and 10 input frames was 108.53 ms. Using the same settings, ReHAR only takes 78.40 ms. Considering that both \cite{msibrahimCVPR16} and \cite{feifei16} predict the activities based on detecting and analyzing every single person and then infer the final activities based on individual actions, they have a similar computation time (4.2 seconds on a GTX 1080 reported in \cite{li2017sbgar}). ReHAR runs an order of magnitude faster than \cite{msibrahimCVPR16} and \cite{feifei16}. Thus, our scheme will be more useful for real-time human activity recognition. 

\begin{table}[h]
  \centering  
  \begin{adjustbox}{width=\linewidth}
  \begin{tabular}{|l|c|c|} 
    \hline
    CNN base net & Time on 10 Frames (ms) & Time on 24 Frames (ms)  \\
    \hline
    VGG16 & 103.65 & 239.04  \\
    InceptionV3 & 78.40 & 192.02 \\
    \hline
  \end{tabular}
  \end{adjustbox}
  \caption{Computation time of ReHAR using different CNN model as its base net (optical flow images generation time included).}
    \label{table:computation_time}
\end{table}

\subsection{Why does our scheme work?}
In previous subsections, we report our comparable results on two well-known activity recognition datasets. In this subsection, we will try to explain the reason why our proposed model works. 

First, we explore the necessity of the LSTM1 and the Global Pooling layers in our scheme by comparing baselines' results on UCFSports dataset. Our proposed model achieves 0.928 mAP (Table \ref{table:comparison_ucfsports}). \textbf{(1)} If we remove LSTM1, stack and feed the output of global layers to a Convolutional layer before ``FC1'' layer, the mAP reduces to 0.766. \textbf{(2)} We only get 0.702 mAP after replacing the LSTM1 with an element-wise sum operation. Baseline (1) and (2) are the best fusion methods discussed in \cite{feichtenhofer2016convolutional}. One can see that our LSTM1 generates much better representations than a simple fusion method. \textbf{(3)} Replacing the Global Pooling layers with flattened layers, the mAP reduces from 0.928 to 0.889. Thus, using Global Pooling layers helps our model achieve a higher accuracy. 

Then, we use the method proposed in \cite{selvaraju2016grad} to compute the gradient of output category with respect to input image. This should tell us how the output category value changes with respect to a small change in input image pixels. We implement this function by modifying the keras-vis toolkit \cite{raghakotkerasvis}, so that the final class-specific information can be passed back through two LSTMs and fully-connected layers. 

\begin{figure*}[h!]
   \centering
   \begin{subfigure}[b]{\textwidth}
      \includegraphics[width=\linewidth]{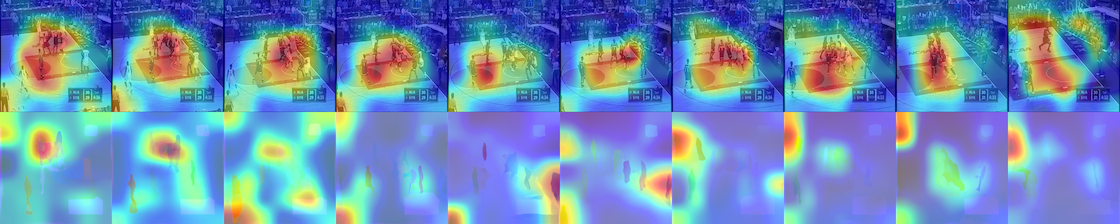}
      \caption{Correctly predict an ``other 2-pointer success'' event on Basketball Dataset. }
      \label{fig:vis1} 
   \end{subfigure}
   \begin{subfigure}[b]{\textwidth}
      \includegraphics[width=\linewidth]{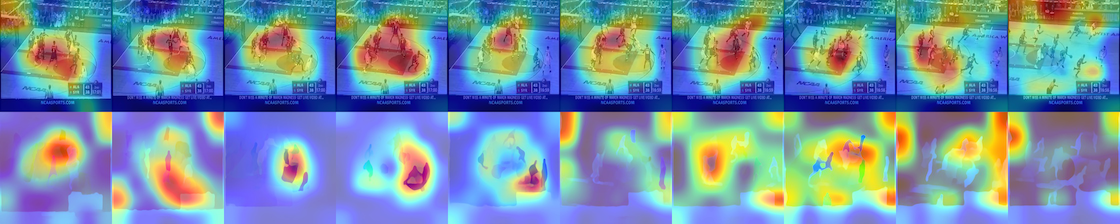}
      \caption{Correctly predict a ``Steal Success'' event on Basketball Dataset.}
      \label{fig:vis2} 
   \end{subfigure}
   \begin{subfigure}[b]{\textwidth}
      \includegraphics[width=\linewidth]{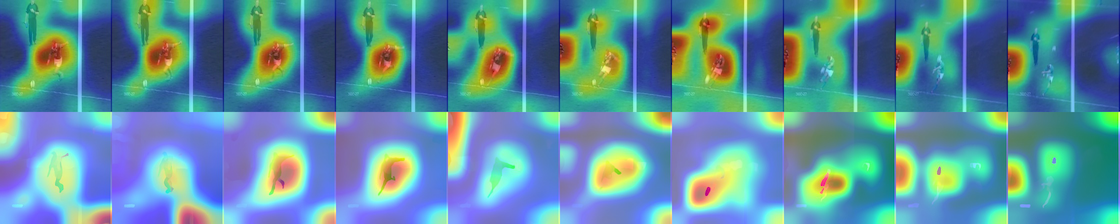}
      \caption{Correctly predict a ``Kicking'' event on UCF Sports Action Dataset.}
      \label{fig:vis3} 
   \end{subfigure}
   \begin{subfigure}[b]{\textwidth}
      \includegraphics[width=\linewidth]{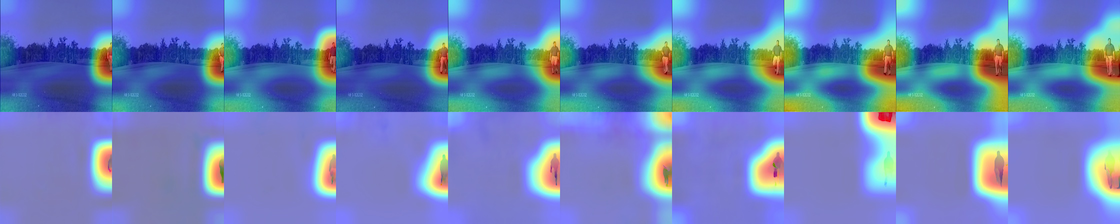}
      \caption{Incorrectly predict a ``Walking'' event as ``Golf'' on UCF Sports Action Dataset.}
      \label{fig:vis4}
   \end{subfigure}
   \caption{Visualized class-specific important regions on input video frames and optical flow images. The top 2 samples are from Basketball Dataset and the bottom 2 are from UCF Sports Dataset. Each sample has two rows of visualized results. The first row shows the results of video frames, while the second row illustrates the results of optical flow images. Because of the limitation of the space, we only visualize 10 frames of each event. }
   \label{fig:visualization}
   \vspace{-4mm}
\end{figure*}

Figure \ref{fig:visualization} shows the visualized results using four samples from the Basketball Dataset and the UCF Sports Action Dataset. \textbf{Figure \ref{fig:vis1}} shows an ``other 2-pointer success'' event and our model correctly predicts it. One can notice that it is hard for a person to find and track the ball through all frames. The visualized result shows that our model focuses on analyzing the players instead of tracking the ball. At the first frame, the group of the players on the video frame draws the attention of our model, while the model pays more attention on the region of the shooter on the optical flow image. At the last frame, the model stares at the region under the net with only one player left. Based on these information as well as features extracted from intermediate frames, the model predicts a correct activity label. \textbf{Figure \ref{fig:vis2}} shows one player successfully steals the ball from another at sixth frame, and then all players are running towards the other side of the basketball court. The model focuses on a larger region of optical flow images after the sixth frame, because all players (including environment) are moving quickly. From \textbf{Figure \ref{fig:vis3}}, one can notice that the model has the capability to detect the key actor from video frames. There are two people on the images and the model highlights the shooter rather than the referee on most frames. In addition, the model highlights the location of the ball at the last 4 optical flow images. All of these prove that our designed model can focus on important and meaningful things. In \textbf{Figure \ref{fig:vis4}}, we visualize a sample that our model wrongly predicts a ``Walking'' event to ``Golf''. The visualized result shows that the model extracts features from the person and the background context on video frames. These features contributes toward ``Golf'' event. We also notice that at the last optical flow image, the model highlights the region of the golf pole which is located between the person's two legs. Maybe these are the reasons why the model has 97\% confidence to label this sample as ``Golf''.

Please refer to our Appendix for more visualized results. 

\section{Conclusion}
\label{sec:conclusion}
In this paper, we propose a novel scheme (ReHAR) to recognize human activities in videos. The proposed model is trainable end-to-end and achieves a higher accuracy than the existing state-of-the-art solutions on both single person activity and group activity datasets. The experimental results also show that ReHAR runs an order of magnitude faster than other schemes. By visualizing the proposed model, we understand what ReHAR learns and notice that it has the potential capability to detect key actors. 

In the near future, we would like to evaluate the impact of using different CNN models, e.g. C3D \cite{tran2014c3d} or MobileNet, as base net. C3D can extract spatiotemporal features from videos, thus using C3D as our base net may achieve higher accuracy. Comparing to other CNN models, MobileNet runs much faster while maintaining accuracy. Thus, using MobileNet as our base net should speed up our model without losing too much accuracy. Moreover, we are planning to explore the performance of our proposed ReHAR on larger datasets, e.g. UCF101 \cite{UCF101} and THUMOS\cite{THUMOS14}, and in other tasks, e.g. activity detection or event proposal generation in untrimmed videos. 
\section{Acknowledgement}
This work is partially supported by a Qualcomm gift and a GPU donated by NVIDIA.

{\small
\bibliographystyle{unsrt}
\bibliography{references}
}

\end{document}


\begin{figure*}[h!]
    \centering
    \begin{subfigure}[b]{\textwidth}
        \includegraphics[width=\linewidth]{../images/app/app_vis7.png}
    \end{subfigure}
    \begin{subfigure}[b]{\textwidth}
        \includegraphics[width=\linewidth]{../images/app/app_vis8.png}
        \caption{Golf}
    \end{subfigure}  
    \begin{subfigure}[b]{\textwidth}
        \includegraphics[width=\linewidth]{../images/app/app_vis13.png}
    \end{subfigure}
    \begin{subfigure}[b]{\textwidth}
        \includegraphics[width=\linewidth]{../images/app/app_vis14.png}
        \caption{\footnotesize Kicking}
    \end{subfigure}
    \begin{subfigure}[b]{\textwidth}
        \includegraphics[width=\linewidth]{../images/app/app_vis16.png}
    \end{subfigure}
    \begin{subfigure}[b]{\textwidth}
        \includegraphics[width=\linewidth]{../images/app/app_vis17.png}
        \caption{\small Lifting}
    \end{subfigure}
\end{figure*}

\begin{figure*}[h!]\ContinuedFloat
    \begin{subfigure}[b]{\textwidth}
        \includegraphics[width=\linewidth]{../images/app/app_vis24.png}
    \end{subfigure}
    \begin{subfigure}[b]{\textwidth}
        \includegraphics[width=\linewidth]{../images/app/app_vis25.png}
        \caption{Run}
    \end{subfigure}
    \begin{subfigure}[b]{\textwidth}
        \includegraphics[width=\linewidth]{../images/app/app_vis37.png}
    \end{subfigure}
    \begin{subfigure}[b]{\textwidth}
        \includegraphics[width=\linewidth]{../images/app/app_vis38.png}
        \caption{Swing-SideAngle}
    \end{subfigure}
    \begin{subfigure}[b]{\textwidth}
        \includegraphics[width=\linewidth]{../images/app/app_vis42.png}
    \end{subfigure}
    \begin{subfigure}[b]{\textwidth}
        \includegraphics[width=\linewidth]{../images/app/app_vis46.png}
        \caption{Walk}
    \end{subfigure}
\end{figure*}